\documentclass{article}
\usepackage[margin=1in]{geometry}

\usepackage{amsmath,amssymb,amsthm}
\usepackage[T1]{fontenc}
\usepackage[utf8]{inputenc}
\usepackage{microtype}
\usepackage{url}
\usepackage{booktabs}
\usepackage{mathtools}
\usepackage{graphicx}
\usepackage{subcaption}
\usepackage{multirow}
\usepackage{algorithm}
\usepackage{algorithmic}
\usepackage{float}
\usepackage{placeins}
\usepackage{xcolor}
\usepackage{listings}
\usepackage{makecell}
\usepackage{tabularx}
\usepackage{balance}
\usepackage{xspace}
\usepackage{hyperref}
\usepackage{cleveref}
\usepackage[numbers]{natbib}

\newcommand{\methodname}{CLEAR}
\DeclareRobustCommand{\method}{%
  \ifmmode
    \mathrm{\methodname}%
  \else
    \textsc{\methodname}\xspace
  \fi
}

\hypersetup{breaklinks=true,colorlinks=true}

\begin{document}

\title{\method: Class-wise Expert Aggregation with Structured Sampling for Long-Tailed Classification}

\author{
Gawon Lim \\
School of Information Sciences \\
University of Illinois Urbana-Champaign \\
\texttt{gawonl2@illinois.edu}
}

\maketitle

\begin{abstract}
Long-tailed classification poses a reliability challenge because models trained on imbalanced data are unevenly reliable across frequent and underrepresented classes. 
While existing methods address imbalance through re-balancing, adjustment, representation learning, or multi-expert modeling, they rarely estimate which expert should be trusted for each class.
This paper proposes \textbf{\method} (\textbf{C}lass-wise re\textbf{L}iability-aware \textbf{E}xpert \textbf{A}ggregation for long-tailed \textbf{R}ecognition), a modular ensemble framework for long-tailed classification.
\method generates diverse experts through threshold-based structured sampling while preserving the full label space, then estimates a class-wise trust score for each expert using a smoothed class-wise precision formulation.
During inference, expert predictions are combined through class-wise generalized product-of-experts aggregation, allowing different experts to be emphasized for different classes.
Experiments on CIFAR-100-LT, ImageNet-LT, and Places-LT across multiple backbones show that \method achieves competitive overall accuracy and particularly strong few-shot performance. 
These results support class-wise expert reliability as a useful design principle for long-tailed ensemble learning.
\end{abstract}

\section{Introduction}

Real-world classification problems rarely follow a balanced label distribution.
In visual recognition, recommendation, entity classification, and knowledge management systems, a small number of frequent classes often dominate the training set, while many rare classes appear only sparsely.
Such long-tailed distributions are especially problematic when rare categories, entities, users, or concepts correspond to high-value or difficult cases where reliable predictions matter most.
As a result, models trained on imbalanced data tend to perform well on majority classes but become unreliable on underrepresented classes.
Thus, the central challenge in long-tailed recognition is not merely low average accuracy, but uneven prediction reliability across different regions of the label space.

Prior work has addressed long-tailed recognition through re-sampling, loss re-weighting, logit adjustment, decoupled classifier learning, contrastive representation learning, and multi-expert modeling.
These methods reduce class-frequency bias, improve feature discrimination, or encourage expert diversity.
However, most existing approaches still model reliability at a coarse level: a classifier is optimized to be globally better, or an expert is assigned a global or sample-dependent role.
In particular, multi-expert methods often encourage diversity or learn sample-level routing, but they typically do not estimate a persistent class-wise reliability profile for each expert.
This leaves an important question underexplored: given multiple experts trained under imbalance, which expert should be trusted for each class?

The proposed framework starts from a simple observation: model reliability in long-tailed classification is inherently class-dependent.
An expert trained close to the original long-tailed distribution may preserve strong many-shot performance, while another expert trained on a more balanced subset may better recognize medium- or few-shot classes.
Therefore, assigning a single global weight to each expert is too restrictive.
To the best of the authors' knowledge, we propose the first long-tailed ensemble framework to place class-wise expert reliability at the center of expert aggregation, rather than relying on globally shared expert weights or secondary calibration signals.

This paper proposes \textbf{\method} (\textbf{C}lass-wise re\textbf{L}iability-aware \textbf{E}xpert \textbf{A}ggregation for long-tailed \textbf{R}ecognition), a modular ensemble framework for long-tailed classification.
\method has two components.
First, \textbf{structured sampling} generates multiple sub-training sets with different imbalance levels while preserving all classes in every subset, producing experts specialized under different class-distribution regimes.
Second, \textbf{class-wise trust-weighted aggregation} assigns each expert a separate reliability score for each class.
The trust score is computed as a smoothed estimate of class-wise precision, measuring how reliably an expert's predictions for each class are supported by reference data. We implement this smoothing using a Beta prior, which yields a closed-form estimate and stabilizes precision when only a few predictions are available.
The final prediction is obtained through a class-wise generalized product-of-experts aggregation, allowing each class to place greater weight on experts estimated to be more reliable for it.

\method is modular: it does not replace existing long-tailed training objectives, but can be combined with components such as Balanced Softmax, Balanced Contrastive Learning, and post-hoc Logit Adjustment. \method is evaluated on CIFAR-100-LT, ImageNet-LT, and Places-LT across multiple backbones.
It achieves competitive overall accuracy and particularly strong few-shot performance on CIFAR-100-LT and Places-LT, reaching 41.25\% and 40.82\% few-shot accuracy, respectively.
On ImageNet-LT, \method remains competitive with 58.43\% overall accuracy, suggesting that class-wise trust weighting generalizes across object-centric and scene-centric long-tailed benchmarks.

The main contributions are summarized as follows:
\begin{itemize}
    \item Class-wise expert reliability is identified as an underexplored principle for long-tailed ensemble learning, showing why global expert weighting is insufficient under severe class imbalance.
    \item \textbf{\method} is introduced as a modular framework that combines structured expert generation with class-wise reliability-aware aggregation, allowing each class to emphasize experts that are more reliable for it.
    \item \method is evaluated across CIFAR-100-LT, ImageNet-LT, and Places-LT, showing competitive overall accuracy and particularly strong few-shot performance across multiple backbone architectures.
\end{itemize}

\begin{figure*}[t]
    \centering
    \includegraphics[width=0.8\textwidth]{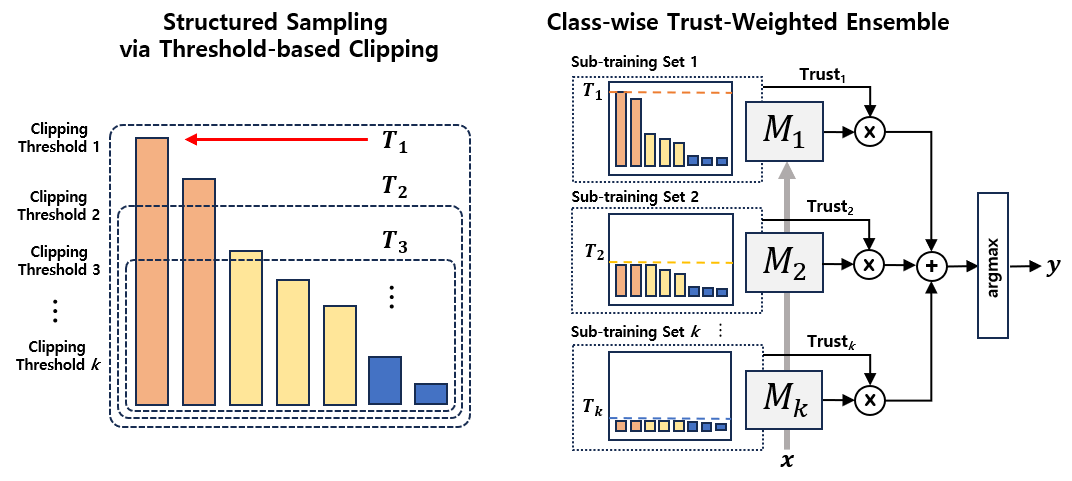}
    \caption{Overview of \method. (Left) Threshold-based structured sampling generates multiple sub-training sets with varying imbalance levels by progressively clipping class frequencies. 
This process produces diverse expert models specialized in different regions of the label distribution. 
(Right) The predictions of these experts are aggregated using class-wise trust weights derived from precision. The final prediction is obtained via trust-weighted aggregation followed by argmax.}
    \label{fig:best-pipeline}
\end{figure*}

\section{Related Work}
\textbf{Re-sampling and imbalance handling.} A classical strategy for multi-class imbalance is the One-vs-All (OVA) decomposition~\cite{rifkin2004ova}, which reduces a multi-class problem into a set of binary classification tasks. While simple and scalable, OVA can suffer under severe class imbalance. To mitigate such imbalance, many methods have explored over-sampling, under-sampling, or hybrid re-sampling strategies. For example, HCBOU~\cite{salehi2025hcbo} utilizes clustering-based over- and under-sampling to preserve local data geometry while reducing class imbalance. Furthermore, EPS~\cite{jabir2023eps} combines partition-based sampling and ensembling under the OVA framework, showing that multiple balanced subsets can improve robustness across classes. These methods demonstrate the value of constructing more balanced training distributions, but they usually focus on generating a single balanced classifier or aggregating classifiers without explicitly modeling class-wise expert reliability.

\textbf{Classifier adjustment and representation learning.} Another major line of work addresses long-tailed recognition by correcting the bias induced by skewed class frequencies. Loss-based methods such as Class-Balanced Loss~\cite{cui2019cbloss} modify the training objective by weighting inversely to the effective number of samples, and LDAM~\cite{cao2019ldam} introduces a label-distribution-aware margin loss to encourage larger margins for minority classes. Standard techniques like Focal Loss~\cite{lin2017focal} apply sample-level control to down-weight well-classified examples dynamically, while DRO-LT~\cite{samuel2021dro} minimizes the worst-case loss over class distribution neighborhoods to ensure distributional robustness. Moreover, decoupled learning~\cite{kang2020decoupling} established an influential paradigm by demonstrating that representation learning and classifier re-balancing can be effectively separated to improve long-tailed performance. Logit-level adjustment methods~\cite{menon2021la} and Balanced Meta-Softmax~\cite{ren2020balanced} also explicitly account for class priors to post-hoc adjust predictions or incorporate them into the loss to accommodate label distribution shifts. Related architectural approaches such as BBN~\cite{zhou2020bbn} use specialized bilateral branches to balance head- and tail-class learning concurrently.

\textbf{Recent long-tailed learning methods.} Recent work has further expanded long-tailed recognition beyond standard loss reweighting and classifier adjustment to emphasize representation discriminability and semantic knowledge. Contrastive learning objectives tailored to imbalanced recognition, such as PaCo~\cite{cui2021paco}, GPaCo~\cite{cui2024gpaco}, and BCL~\cite{zhu2022bcl}, have been introduced to improve feature separation and form a balanced feature space through class-center embeddings. ACL~\cite{ma2025acl} analyzes gradient conflicts in supervised contrastive learning under long-tailed distributions and proposes an aligned contrastive objective to reduce imbalanced attraction and repulsion effects. Furthermore, LTRL~\cite{zhao2024ltrl} introduces a reflective learning paradigm that dynamically reviews past predictions and corrects gradient conflicts during training. Additionally, ConCutMix~\cite{pan2024concutmix} enhances tail class representation by explicitly constructing semantically consistent labels for augmented mixed samples, and MGS~\cite{liu2025mgs} exploits multi-granularity semantic information leveraging large language models to transfer knowledge dynamically from head classes to tail classes. These recent approaches suggest that effective long-tailed recognition requires not only distribution correction but also better use of semantic structure and training dynamics.

\textbf{Multi-expert and ensemble methods.} Multi-expert learning has become a strong direction for long-tailed recognition. RIDE~\cite{wang2021ride} trains multiple diverse experts within a shared architecture and encourages expert diversity to improve robustness across the label space. SADE~\cite{zhang2022sade} extends this direction by introducing self-supervised aggregation of diverse experts for test-agnostic long-tailed recognition. MDCS~\cite{zhao2023mdcs} further improves multi-expert learning by promoting expert diversity and reducing model variance through consistency self-distillation. Knowledge distillation has also been explored for long-tailed learning; for example, Balanced Knowledge Distillation~\cite{zhang2023bkd} transfers balanced knowledge from teacher models to improve performance under imbalanced class distributions. Furthermore, recent studies like MORE~\cite{luo2025more} mitigate imbalance by directly rebalancing the model's parameter space using a low-rank parameter component to reserve dedicated capacity for minority classes. These methods show that diversity, aggregation, knowledge transfer, and structural re-allocation are important for long-tailed recognition. However, existing multi-expert methods typically rely on shared-backbone expert diversification, sample-level routing, consistency regularization, or global/adaptive expert weighting. They do not directly estimate how reliable each expert is for each individual class.

\textbf{Positioning of \method.} \method is most closely related to multi-expert and ensemble-based long-tailed recognition, but it differs from prior work in two important ways. First, instead of generating experts only through architectural branches or shared-backbone specialization, \method constructs independent experts from structured sub-training sets that represent different imbalance regimes. This allows each expert to learn from a different view of the original long-tailed distribution. Second, rather than assigning a single global importance score to each model or routing samples to experts, \method estimates class-wise trust for each expert and normalizes reliability across experts separately for every class. This class-dependent aggregation is particularly suitable for long-tailed settings, where the most reliable expert for head classes may not be the most reliable expert for tail classes.

\section{Method}

We propose \textbf{\method}, a framework for imbalanced multi-class classification based on \textbf{class-wise reliability-aware expert aggregation}.
Unlike conventional ensemble methods that assign a single global weight per model, \method performs class-dependent weighting of expert predictions, enabling different experts to contribute more strongly to different classes as shown in Fig.~\ref{fig:best-pipeline}.

\method is built upon two key components:
\begin{itemize}
    \item \textbf{Diverse expert generation}: constructing multiple expert models that specialize in different class imbalance regimes via structured sampling;
    \item \textbf{Trust-weighted aggregation}: combining expert predictions using class-wise trust scores derived from smoothed class-wise precision.
\end{itemize}

The key idea of \method is that model reliability is inherently \textbf{class-dependent} in long-tailed settings. 
By estimating a separate trust score for each model and class, \method adaptively weights expert contributions so that more reliable models have greater influence for specific classes.

We first describe the trust-based expert aggregation mechanism, and then introduce structured sampling to induce expert diversity.

\subsection{Class-wise Trust Estimation}

Let $\mathcal{D}=\{(x_i,y_i)\}_{i=1}^N$ denote a classification dataset with $C$ classes.
Assume that we are given $M$ pretrained models (experts) and
different models exhibit heterogeneous performance across classes in class-imbalanced settings.
This implies that a model that performs well on majority classes may not be reliable on minority classes, while another model may specialize in a small subset of difficult classes.
Therefore, instead of assigning a single global weight to each model, we devise a class-wise ensemble mechanism that captures per-class specialization.

For each model $m$ and class $c$, we define the class-wise reliability parameter
\begin{equation}
\theta_{m,c} = P(\text{correct} \mid \hat{y}_m=c),
\end{equation}
which represents the probability that expert $m$ is correct when it predicts class $c$.

Directly estimating $\theta_{m,c}$ from empirical precision can be unstable when only a small number of predictions are available for class $c$. To stabilize this estimate, we apply \textbf{Beta-prior smoothing}:
\begin{equation}
\theta_{m,c} \sim \mathrm{Beta}(\alpha_0, \beta_0),
\end{equation}
and model the statistical observations of model performance as
\begin{equation}
n_{m,c} \sim \mathrm{Binomial}(N_{m,c}, \theta_{m,c}),
\end{equation}
where $n_{m,c}$ is the number of correct predictions (true positives) and $N_{m,c}$ is the total number of predictions for class $c$ (true positives + false positives).

Then, the posterior distribution is given by
\begin{equation}
\theta_{m,c} \mid \mathcal{D} \sim \mathrm{Beta}(\alpha_0 + n_{m,c}, \beta_0 + N_{m,c} - n_{m,c}),
\end{equation}
and we define the class-wise trust score as the posterior mean as follows:
\begin{equation}
q_{m,c} = \mathbb{E}[\theta_{m,c} \mid \mathcal{D}] 
= \frac{\alpha_0 + n_{m,c}}{\alpha_0 + \beta_0 + N_{m,c}}. \label{eq:q_mc}
\end{equation}
Without smoothing, the corresponding empirical estimate reduces to
$q_{m,c} = \frac{n_{m,c}}{N_{m,c}}$. This formulation provides a smoothed estimate of class-wise precision with two desirable properties.
First, it smooths unreliable estimates when $N_{m,c}$ is small.
Second, it avoids overconfident trust values for rare classes, where raw empirical precision may be highly unstable.

We normalize the class-wise trust scores in (\ref{eq:q_mc}) across experts:
\begin{equation}
w_{m,c} =
\frac{\exp(\tau q_{m,c})}
{\sum_{j=1}^{M} \exp(\tau q_{j,c})}, \label{eq:w_mc}
\end{equation}
where $\tau \geq 0$ is a sharpness parameter that controls how strongly the class-wise weighting concentrates on the most reliable model for class $c$.

Note that 
$\sum_{m=1}^{M} w_{m,c}=1$ for every class $c$, so $w_{m,c}$ can be interpreted as the relative contribution of the expert model $m$ to class $c$ among all experts.

This class-wise normalization is important in long-tailed recognition, where the expert that performs best for one class is often not the best for another. As a result, the ensemble can adaptively favor different experts for different classes.

\subsection{Class-wise Trust-Weighted Expert Aggregation}

We treat each model as a noisy expert whose reliability depends on the class.
The latent reliability variable $\theta_{m,c}$ determines the strength of evidence contributed by model $m$ for class $c$.

Under this view, the likelihood contribution of model $m$ for class $c$ is proportional to
\begin{equation}
p_m(c\mid x)^{\theta_{m,c}}.
\end{equation}
This can be viewed as a class-dependent generalized Product-of-Experts (gPoE), where each expert contributes multiplicatively with a class-dependent exponent.

Motivated by a conditional-independence approximation commonly used in product-of-experts aggregation, the posterior over the label is proportional to
\begin{equation}
p(y=c\mid x)
\propto
\prod_{m=1}^{M} p_m(c\mid x)^{\theta_{m,c}}.
\end{equation}
Because experts are trained on overlapping data and related architectures, this conditional-independence assumption should be understood as an approximation that motivates a practical aggregation rule rather than as an exact generative model.

We first approximate $\theta_{m,c}$ by its posterior mean $q_{m,c}$ in (\ref{eq:q_mc}), and then obtain normalized weights $w_{m,c}$ via a softmax transformation in (\ref{eq:w_mc}).
Given an input $x$, each expert $m$ produces a predictive distribution $p_m(c \mid x)$.
By replacing $\theta_{m,c}$ with the normalized trust $w_{m,c}$, 
\method aggregates expert predictions using a class-wise gPoE formulation:
\begin{equation}
p_{\mathrm{\method}}(c \mid x)
\propto
\prod_{m=1}^{M} p_m(c \mid x)^{w_{m,c}}.
\end{equation}

Equivalently, in log-space,
\begin{equation}
S_c(x) = \sum_{m=1}^{M} w_{m,c} \log p_m(c \mid x), \label{eq:S_c}
\end{equation}
and the final predictive distribution is obtained via softmax:
\begin{equation}
p_{\mathrm{\method}}(c \mid x) =
\frac{\exp(S_c(x))}
{\sum_{k=1}^{C} \exp(S_k(x))}. \label{eq:p_method}
\end{equation}
The final predicted class is then obtained by
\begin{equation}
\hat{y} = \arg\max_c p_{\mathrm{\method}}(c \mid x).
\end{equation}

This formulation enables \textbf{class-wise generalized Product-of-Experts aggregation}: for each class, experts with higher trust scores contribute more strongly, allowing different experts to specialize across classes. This fundamentally differs from conventional ensembles, which use globally shared weights and cannot capture class-dependent expert reliability.

In particular, the sharpness parameter $\tau$ in (\ref{eq:w_mc})  interpolates between uniform aggregation and hard class-wise expert selection.
As $\tau\to 0$, the weights satisfy $w_{m,c}\to 1/M$, yielding
\begin{equation}
S_c(x)\to \frac{1}{M}\sum_{m=1}^{M}\log p_m(c\mid x),
\end{equation}
which is equivalent to the geometric mean of expert predictions.
In contrast, as $\tau\to\infty$, if the maximizer 
\(
m_c^*=\arg\max_m q_{m,c}
\)
is unique, then $w_{m,c}\to \mathbf{1}[m=m_c^*]$, and the aggregation reduces to the single most reliable expert for class $c$.
Therefore, $\tau$ controls the trade-off between soft cooperation among experts and hard class-wise specialization.

\subsection{Structured Sampling for Expert Generation}

The effectiveness of the class-wise trust-weighted ensemble depends on the diversity and specialization of the expert models.
To induce such diversity, \method introduces \textbf{structured sampling}, which generates multiple sub-training sets with varying imbalance levels and encourages different models to specialize in different regions of the label space.

\method constructs experts over multiple stages, where each stage introduces a new sub-training set and a corresponding expert model.
Each sub-training set is obtained by threshold-based clipping of the original training data: classes with instance counts larger than a clipping threshold are downsampled, while classes with equal or fewer samples are fully retained.

Let $n_c$ denote the number of training samples in class $c$, and let $\mathcal{D}_c$ denote the set of samples belonging to class $c$.
At stage $i$, a clipping threshold $T_i$ is defined, and the corresponding sub-training set $\mathcal{S}_i$ is constructed as
\begin{equation}
\mathcal{S}_i
=
\bigcup_{c=1}^{C}
\mathrm{Sample}(\mathcal{D}_c, \min(n_c, T_i)),
\label{eq:structured_sampling}
\end{equation}
where $\mathrm{Sample}(\mathcal{D}_c, k)$ denotes uniformly sampling $k$ examples from class $c$ without replacement.
Thus, classes with more than $T_i$ samples are downsampled to $T_i$, while classes with fewer than $T_i$ samples are fully retained.

By progressively decreasing the threshold across stages, \method generates a sequence of sub-training sets with varying imbalance ratios while preserving all classes.
Early stages remain close to the original long-tailed distribution, whereas later stages become increasingly balanced.
Training one model on each subset yields a collection of experts $\{p_i(c \mid x)\}_{i=1}^{M}$ with heterogeneous class-wise behavior, since different imbalance regimes favor different classes.

Importantly, this diversity is structured rather than random: each expert is exposed to a systematically modified class distribution, encouraging complementary specialization across classes.
This property is particularly important for \method, because the class-wise trust-weighted aggregation is most effective when different experts are reliable on different regions of the label space.

We consider structured sampling with \textbf{Exponential Decay Clipping
(EDC)}. At stage $i$, EDC defines the clipping threshold as
\begin{equation}
T_i = \left\lfloor C_{\max} ~\delta^{\,i-1} \right\rfloor,
\label{eq:threshold_edc}
\end{equation}
where $C_{\max} = \max_c n_c$ is the maximum class frequency in the original
training set and $\delta \in (0,1)$ controls the decay rate. This schedule
produces a sequence of sub-training sets that gradually moves from the
original long-tailed distribution toward more balanced distributions while
retaining all classes at every stage.

Note that \method is not restricted to EDC: other schedules, such as uniform intervals,
quantile-based thresholds, or clustering-based thresholds, can also be used
to instantiate the structured sampling step. 

Each expert is trained on the subset $\mathcal{S}_i$ in (\ref{eq:structured_sampling}).
For classes with $n_c > T_i$, threshold-based clipping leaves samples that are not used to train expert $M_i$; these held-out samples are used to estimate the corresponding class-wise prediction statistics. For classes with $n_c \leq T_i$, all available samples are included in $\mathcal{S}_i$, and therefore no held-out samples remain. In this case, the class-wise statistics are estimated from training-side (in-bag) predictions on $\mathcal{S}_i$. The resulting confusion statistics are then used to compute the class-wise trust score $q_{i,c}$. The in-bag estimate for classes with $n_c \leq T_i$ may exhibit optimistic bias, but it avoids further reducing the already limited number of samples available for rare classes.

Overall, structured sampling provides a principled mechanism for generating complementary experts whose class-wise reliabilities are later exploited by the trust-weighted aggregation.
Algorithm~\ref{alg:method} summarizes the full training and inference pipeline.

\begin{algorithm}[t]
\caption{\method: Class-wise Trust-Weighted Ensemble with Structured Sampling}
\label{alg:method}
\begin{algorithmic}[1]
\REQUIRE Training dataset $\mathcal{D}_{\text{train}}$, test dataset $\mathcal{D}_{\text{test}}$, number of experts $M$, base architecture $\mathcal{M}$, prior hyperparameters $\alpha_0,\beta_0$, sharpness parameter $\tau$
\ENSURE Final prediction $\hat{y}$ for each $x \in \mathcal{D}_{\text{test}}$

\STATE Compute class counts $n_c$ for each class $c$ in $\mathcal{D}_{\text{train}}$
\FOR{$m = 1$ to $M$}
    \STATE Define threshold $T_m$ according to the chosen schedule (e.g., (\ref{eq:threshold_edc}))
    \STATE Construct sub-training set $\mathcal{S}_m$ by (\ref{eq:structured_sampling})
    \STATE Train expert $M_m \gets \mathrm{Train}(\mathcal{M}, \mathcal{S}_m)$
    \STATE For classes with $n_c > T_m$, compute class-wise counts $n_{m,c}$ and $N_{m,c}$ from held-out samples excluded from $\mathcal{S}_m$ by threshold-based clipping
    \STATE For classes at or below $T_m$, where all samples are used for training and no held-out samples remain, compute class-wise counts $n_{m,c}$ and $N_{m,c}$ from training-side (in-bag) predictions within $\mathcal{S}_m$
    \STATE Estimate class-wise trust scores $q_{m,c}$ by (\ref{eq:q_mc})
\ENDFOR

\FOR{each class $c$}
    \STATE Compute normalized class-wise weights $w_{m,c}$ across experts by (\ref{eq:w_mc})
\ENDFOR

\FOR{each $x \in \mathcal{D}_{\text{test}}$}
    \FOR{each class $c$}
        \STATE Compute aggregation score $S_c(x)$ by (\ref{eq:S_c})
    \ENDFOR
    \STATE Compute final predictive distribution $p_{\mathrm{\method}}(c \mid x)$ by (\ref{eq:p_method})
    \STATE Predict $\hat{y} \gets \arg\max_c p_{\mathrm{\method}}(c \mid x)$
\ENDFOR
\end{algorithmic}
\end{algorithm}

\section{Experiments}

\subsection{Experimental Setup}

\textbf{Datasets.} We evaluate our proposed \method framework on three widely used long-tailed recognition benchmarks: CIFAR-100 Long-Tailed~\cite{cao2019ldam}, ImageNet-LT~\cite{liu2019large}, and Places-LT~\cite{liu2019large}. 
\begin{itemize}
    \item \textbf{CIFAR-100-LT:} The original CIFAR-100 contains 50,000 training images and 10,000 test images across 100 classes. We use the long-tailed version constructed by exponentially decaying the number of training samples per class. In our experiments, we focus on the severe imbalance ratio of 100.
    \item \textbf{ImageNet-LT:} This dataset is a long-tailed subset sampled from the large-scale ImageNet-2012 dataset following a Pareto distribution. It contains 115.8K training images from 1,000 categories, with the number of images per class ranging from 5 to 1,280.
    \item \textbf{Places-LT:} Constructed from the large-scale Places365-Standard dataset, this benchmark contains images from 365 scene categories, with class cardinality ranging from 5 to 4,980.
\end{itemize}

\textbf{Evaluation Metrics.} Following standard evaluation protocols in long-tailed visual recognition, we report the top-1 accuracy on the balanced test sets. To provide a comprehensive analysis of performance across different class frequencies, we further report accuracy on three sub-splits: Many-shot ($>$100 training images), Medium-shot (20$\sim$100 training images), and Few-shot ($<$20 training images).

\textbf{Implementation Details.} We employ backbone architectures commonly used for each benchmark. For CIFAR-100-LT, we evaluate \method with ResNet-32, WRN-28-10, and ViT-B/16. For ImageNet-LT, we adopt ResNeXt-50, and for Places-LT, we use ImageNet-pretrained ResNet-152. For the \method framework, we use EDC-based structured sampling with dataset-specific decay rates. All reported \method results are averaged over five random seeds, 40--44.

In all reported experiments, class-wise trust is estimated using held-out samples excluded by threshold-based clipping when $n_c > T_m$. For classes with $n_c \leq T_m$, all samples are included in the corresponding sub-training set $\mathcal{S}_m$, so the class-wise statistics are estimated from training-side (in-bag) predictions. The resulting prediction statistics are used to compute $q_{m,c}$ with $\alpha_0=\beta_0=1$.

The dataset-specific model configurations are summarized in Table~\ref{tab:dataset_model_config}.

\begin{table}[t]
\centering
\caption{Dataset and model configuration summary.}
\label{tab:dataset_model_config}
\small
\renewcommand{\arraystretch}{1.05}
\begin{tabular}{l l c c l c c}
\toprule
\textbf{Dataset} & \textbf{Model} & \textbf{Ep.} & \textbf{BS}
& \textbf{Opt.} & \textbf{Init LR} & \textbf{WD} \\
\midrule
CIFAR-100-LT & ResNet-32  & 200 & 128 & SGD$^a$   & 0.1        & 5e-4 \\
CIFAR-100-LT & WRN-28-10  & 200 & 128 & SGD$^a$   & 0.1        & 5e-4 \\
CIFAR-100-LT & ViT-B/16   &  50 & 128 & AdamW$^b$ & 1e-5/1e-3  & 0.05 \\
ImageNet-LT  & ResNeXt-50 & 180 &  64 & SGD$^c$   & 0.025      & 5e-4 \\
Places-LT    & ResNet-152 &  30 & 128 & SGD$^d$   & 0.001/0.01 & 4e-4 \\
\bottomrule
\end{tabular}

\vspace{0.35em}
\begin{minipage}{\columnwidth}
\raggedright
\footnotesize
$^a$ Nesterov momentum=0.9; WarmupMultiStep scheduler with milestones at 160/180 epochs, $\gamma=0.1$, and 5 warmup epochs.\\
$^b$ Backbone/head split learning rates; WarmupCosine scheduler with 2 warmup epochs.\\
$^c$ Nesterov momentum=0.9; WarmupCosine scheduler with 5 warmup epochs.\\
$^d$ Nesterov momentum=0.9; backbone/fc split learning rates; Warmup$\to$LinearDecay scheduler with 5 warmup epochs.
\end{minipage}
\end{table}

\textbf{Training Objectives and Inference Adjustments.}
Unless otherwise specified, each \method expert is trained with Balanced Softmax (BSM)~\cite{ren2020balanced}, which corrects the standard softmax objective by incorporating class-frequency information into the logits. 
Given the logit $f_c(x)$ for class $c$ and the number of training samples $n_c$, BSM modifies the normalized probability as
\begin{equation}
p_{\mathrm{BSM}}(y=c \mid x)
=
\frac{n_c \exp(f_c(x))}
{\sum_{j} n_j \exp(f_j(x))}.
\end{equation}
This objective encourages the model to account for the long-tailed label distribution during training while preserving the standard cross-entropy form.

We further consider two optional extensions. 
First, we use Balanced Contrastive Learning (BCL)~\cite{zhu2022bcl} as an auxiliary representation-learning loss:
\begin{equation}
\mathcal{L}
=
\mathcal{L}_{\mathrm{BSM}}
+
\lambda_{\mathrm{BCL}}\mathcal{L}_{\mathrm{BCL}}.
\end{equation}
Here, $\mathcal{L}_{\mathrm{BCL}}$ is applied to $\ell_2$-normalized projected features using a lightweight projection head, which is discarded at inference time. 
This extension is intended to improve feature discrimination, especially for medium- and few-shot classes, without changing the \method aggregation procedure.

Second, we evaluate post-hoc Logit Adjustment (LA)~\cite{menon2021la} at inference time. 
For each expert, the adjusted logit is computed as
\begin{equation}
\tilde{f}_c(x)
=
f_c(x)
-
\alpha_{\mathrm{LA}} \log n_c,
\end{equation}
where $\alpha_{\mathrm{LA}}$ controls the strength of the correction. 
Since BSM already incorporates class-frequency information during training, we use LA only as a lightweight inference-time calibration option and tune $\alpha_{\mathrm{LA}}$ conservatively. 
Importantly, both BCL and LA are modular additions: BCL affects expert training, LA affects inference logits, and neither changes the proposed class-wise trust estimation or \method ensemble aggregation.

\subsection{Comparison with State-of-the-Art Methods}

We compare \method against a broad range of state-of-the-art long-tailed recognition methods, including loss re-balancing methods, decoupled learning baselines, contrastive learning approaches, and multi-expert ensemble methods. 
Rather than treating long-tailed recognition as a single-axis optimization problem, we analyze performance across many-, medium-, and few-shot groups, since different algorithms often show different strengths depending on class frequency. 
In this context, \method is designed not as a single loss re-weighting method, but as a modular ensemble framework that can incorporate complementary long-tailed learning components such as BSM, BCL, and post-hoc Logit Adjustment (LA).

\begin{table}[t]
\centering
\small
\setlength{\tabcolsep}{4pt}
\caption{Top-1 accuracy (\%) comparison on CIFAR-100-LT (Imbalance Ratio = 100) using ResNet-32.}
\label{tab:cifar100lt_sota}
\begin{tabular}{lcccc}
\toprule
\textbf{Method}
& \textbf{Many} & \textbf{Medium} & \textbf{Few} & \textbf{All} \\
\midrule
\multicolumn{5}{l}{\textit{Standard long-tailed baselines}} \\
\midrule
Softmax (CE)~\cite{luo2025more}
& 73.1 & 45.1 & 9.2 & 44.1 \\

$\tau$-norm~\cite{kang2020decoupling}
& 61.4 & 42.5 & 15.7 & 41.4 \\

LADE~\cite{hong2021disentangling}
& -- & -- & -- & 45.4 \\

MiSLAS~\cite{zhong2021calibration}
& -- & -- & -- & 47.0 \\

DRO-LT~\cite{samuel2021dro}
& 64.7 & 50.0 & 23.8 & 47.3 \\

Logit Adjustment (LA)~\cite{menon2021la}
& 65.3 & 51.7 & 31.9 & 50.5 \\

Balanced Softmax~\cite{ren2020balanced}
& -- & -- & -- & 50.8 \\

\midrule
\multicolumn{5}{l}{\textit{Multi-expert and recent methods}} \\
\midrule
RIDE (3 experts)~\cite{wang2021ride}
& 68.1 & 49.2 & 23.9 & 48.0 \\

SADE~\cite{zhang2022sade}
& 61.6 & 50.5 & 33.9 & 49.4 \\

LA + MORE~\cite{luo2025more}
& 65.3 & 52.3 & 33.6 & 51.2 \\

BCL~\cite{zhu2022bcl}
& 67.2 & 53.1 & 32.9 & 51.9 \\

ACL~\cite{ma2025acl}
& -- & -- & -- & 52.6 \\

ConCutMix~\cite{pan2024concutmix}
& 67.4 & 53.9 & 35.8 & 53.2 \\

MGS~\cite{liu2025mgs}
& 68.3 & 54.1 & 37.2 & 54.0 \\

\midrule
\textbf{\method} (BSM)
& 68.54 & 51.03 & 37.45 & 53.09 {\scriptsize $\pm$0.77} \\

\textbf{\method} (BSM/BCL)
& 67.68 & 50.86 & 39.32 & 53.28 {\scriptsize $\pm$0.76} \\

\textbf{\method} (BSM/LA)
& 67.20 & 50.66 & 39.35 & 53.06 {\scriptsize $\pm$0.84} \\

\textbf{\method} (BSM/BCL/LA)
& 66.39 & 50.46 & 41.25 & 53.24 {\scriptsize $\pm$0.84} \\

\midrule
\multicolumn{5}{l}{\textit{Stronger augmentation / alternate settings}} \\
\midrule
PaCo$^\dagger$~\cite{cui2021paco}
& -- & -- & -- & 52.0 \\

GPaCo$^\dagger$~\cite{cui2024gpaco}
& -- & -- & -- & 52.3 \\

BCL$^\dagger$~\cite{zhu2022bcl}
& 69.7 & 53.8 & 35.5 & 53.9 \\

MDCS$^\dagger$~\cite{zhao2023mdcs}
& 72.4 & 57.8 & 35.0 & 56.1 \\

\bottomrule
\end{tabular}
\vspace{0.3em}
\begin{flushleft}
\footnotesize
$^\dagger$ denotes methods trained with stronger augmentation (e.g., RandAugment) or longer training schedules (400 epochs).
Dashes indicate that shot-wise results were not reported in the corresponding paper.
Values after $\pm$ denote the standard deviation across five runs with different random seeds.
\method uses $\tau=2.0$, decay rate $\delta=0.95$, and 15 stages with $\alpha_{\mathrm{LA}}=0.1$ and $\lambda_{\mathrm{BCL}}=0.1$.
\end{flushleft}
\end{table}

\textbf{Results on CIFAR-100-LT.} 
As shown in Table~\ref{tab:cifar100lt_sota}, \method achieves highly competitive overall performance while showing particularly strong robustness on tail classes. 
The unified variant \method (BSM/BCL/LA) achieves a Few-shot accuracy of 41.25\%, outperforming strong recent baselines such as SADE (33.9\%), ConCutMix (35.8\%), and MGS (37.2\%) under the standard 200-epoch training schedule. 
This result is important because the Few-shot split is typically the most difficult regime in long-tailed recognition, where models often overfit to head classes and fail to learn sufficiently discriminative minority-class representations. 
\method also remains competitive with methods trained using much longer schedules and stronger augmentations, such as BCL$^\dagger$ and PaCo$^\dagger$. 
These results suggest that structured sampling and class-wise trust-weighted aggregation can complement existing long-tailed training objectives by improving minority-class recognition while maintaining competitive overall performance.

\begin{table}[t]
\centering
\caption{Top-1 accuracy (\%) of \method on CIFAR-100-LT (Imbalance Ratio = 100) using different backbones.}
\label{tab:cifar100lt_la_backbone}
\small
\setlength{\tabcolsep}{4pt}
\begin{tabular}{llcccc}
\toprule
\textbf{Backbone} & \textbf{Config} & \textbf{Many} & \textbf{Med.} & \textbf{Few} & \textbf{All} \\
\midrule
\multirow{2}{*}{ResNet-32}
& BSM             & 67.65 & 51.61 & 36.53 & 52.70 {\scriptsize$\pm$0.97} \\
& BSM/LA          & 66.33 & 51.38 & 38.45 & 52.73 {\scriptsize$\pm$0.96} \\
\midrule
\multirow{2}{*}{WRN-28-10}
& BSM             & 76.82 & 56.45 & 30.07 & 55.67 {\scriptsize$\pm$0.47} \\
& BSM/LA          & 76.26 & 56.67 & 32.18 & 56.18 {\scriptsize$\pm$0.51} \\
\midrule
\multirow{2}{*}{ViT-B/16}
& BSM             & 94.77 & 88.48 & 82.05 & 88.75 {\scriptsize$\pm$0.31} \\
& BSM/LA          & 94.61 & 88.68 & 82.95 & 89.04 {\scriptsize$\pm$0.28} \\
\bottomrule
\end{tabular}

\vspace{0.3em}
\begin{minipage}{0.98\linewidth}
\footnotesize
\textit{Note.} All results are averaged over five random seeds.
\method uses $\tau=2.0$, decay rate $\delta=0.9$, and 15 stages.
BSM/LA denotes post-hoc Logit Adjustment with $\alpha_{\mathrm{LA}}=0.1$ applied at inference time.
\end{minipage}
\end{table}

Table~\ref{tab:cifar100lt_la_backbone} further analyzes the effect of post-hoc Logit Adjustment (LA) across different backbone architectures on CIFAR-100-LT. 
Across ResNet-32, WRN-28-10, and ViT-B/16, LA consistently improves Few-shot accuracy by +1.92, +2.11, and +0.90 percentage points, respectively. 
The overall accuracy also improves for all backbones, although the gain is modest compared with the Few-shot improvement. 
This suggests that LA primarily serves as a lightweight tail-class calibration mechanism: it slightly suppresses head-class predictions, leading to a small reduction in Many-shot accuracy, but improves minority-class recognition enough to preserve or improve overall accuracy. 
The largest gain is observed for WRN-28-10, whose strong Many-shot accuracy before LA indicates a larger residual head-class bias. 
These results support the modularity of \method, showing that inference-time calibration can be added without retraining the experts while improving tail robustness.

\textbf{Results on ImageNet-LT.} 
Table~\ref{tab:imagenetlt_sota} summarizes the comparison on the large-scale ImageNet-LT dataset using the ResNeXt-50 backbone. 
\method remains competitive across class-frequency groups and provides strong tail-class performance under the evaluated setting. 
The fully integrated \method (BSM/BCL/LA) variant achieves an overall accuracy of 58.43\% under the training schedule used in our experiments, outperforming conventional re-balancing methods such as Balanced Softmax (51.4\%) and decoupled learning baselines such as cRT (49.6\%) and LWS (49.9\%). 
Although some methods achieve higher accuracy on particular shot groups, \method remains competitive as a modular ensemble framework that combines experts through class-wise reliability. 
In particular, \method (BSM/LA) attains 43.48\% accuracy on the Few-shot split, remaining competitive with SADE (43.5\%) and outperforming several contrastive and ensemble baselines such as BCL (36.6\%), ACL (40.6\%), and RIDE (35.1\%). 
This indicates that post-hoc calibration can still provide meaningful tail-class gains even after balanced expert training.

\begin{table}[t]
\centering
\small
\setlength{\tabcolsep}{4pt}
\caption{Top-1 accuracy (\%) comparison on ImageNet-LT using ResNeXt-50.}
\label{tab:imagenetlt_sota}
\begin{tabular}{lcccc}
\toprule
\textbf{Method} & \textbf{Many} & \textbf{Medium} & \textbf{Few} & \textbf{All} \\
\midrule
\multicolumn{5}{l}{\textit{Standard long-tailed baselines}} \\
\midrule
Softmax (CE) & 65.9 & 37.5 & 7.7 & 44.4 \\
$\tau$-norm~\cite{kang2020decoupling} & 59.1 & 46.9 & 30.7 & 49.4 \\
cRT~\cite{kang2020decoupling} & 61.8 & 46.2 & 27.4 & 49.6 \\
LWS~\cite{kang2020decoupling} & 60.2 & 47.2 & 30.3 & 49.9 \\
Balanced Softmax~\cite{ren2020balanced} & 62.2 & 48.8 & 29.8 & 51.4 \\
Causal Norm~\cite{tang2020tde} & 62.7 & 48.8 & 31.6 & 51.8 \\
LADE~\cite{hong2021disentangling} & 62.3 & 49.3 & 31.2 & 51.9 \\
MiSLAS~\cite{zhong2021calibration} & 62.0 & 49.1 & 32.8 & 51.4 \\
DisAlign~\cite{zhang2021disalign} & 62.7 & 52.1 & 31.4 & 53.4 \\
\midrule
\multicolumn{5}{l}{\textit{Multi-expert and recent methods}} \\
\midrule
PaCo~\cite{cui2021paco} & 63.2 & 51.6 & 39.2 & 54.4 \\
RIDE~\cite{wang2021ride} & 68.0 & 52.9 & 35.1 & 56.3 \\
BCL~\cite{zhu2022bcl} & 67.9 & 54.2 & 36.6 & 57.1 \\
SADE~\cite{zhang2022sade} & 66.5 & 57.0 & 43.5 & 58.8 \\
SADE+RL~\cite{zhang2022sade,zhao2024ltrl} & 66.3 & 58.3 & 47.8 & 60.2 \\
ConCutMix~\cite{pan2024concutmix} & 70.7 & 56.6 & 39.8 & 59.7 \\
MGS/SKCL~\cite{liu2025mgs} $^\ast$ & 71.5 & 56.8 & 41.5 & 60.4 \\
ACL~\cite{ma2025acl} & 70.7 & 59.1 & 40.6 & 61.1 \\
\midrule
\textbf{\method} (BSM) & 68.55 & 55.16 & 39.36 & 58.25  {\scriptsize $\pm$0.06} \\
\textbf{\method} (BSM/BCL) & 68.19 & 55.74 & 39.48 & 58.40  {\scriptsize $\pm$0.02} \\
\textbf{\method} (BSM/LA) & 67.01 & 55.49 & {43.48} & 58.36  {\scriptsize $\pm$0.07} \\
\textbf{\method} (BSM/BCL/LA) & 66.57 & 56.03 & 43.39 & {58.43}  {\scriptsize $\pm$0.06} \\
\midrule
\multicolumn{5}{l}{\textit{Stronger augmentation / alternate settings}} \\
\midrule
PaCo$^\dagger$~\cite{cui2021paco} & 67.5 & 56.9 & 36.7 & 58.2 \\
ConCutMix$^\dagger$~\cite{pan2024concutmix} & 72.1 & 58.4 & 40.8 & 61.3 \\
SADE+RL$^\dagger$~\cite{zhang2022sade,zhao2024ltrl} & 67.9 & 61.2 & 47.8 & 62.0 \\
MDCS+RL$^\dagger$~\cite{zhao2023mdcs,zhao2024ltrl} & 72.7 & 59.5 & 46.0 & 62.7 \\
\bottomrule
\end{tabular}
\vspace{0.3em}
\begin{flushleft}
\footnotesize
$^\ast$ MGS/SKCL reports results using a ResNet-50 backbone.
$^\dagger$ denotes methods trained with stronger augmentation or alternate settings, such as RandAugment and longer training schedules.

\method uses $\tau=2.0$, decay rate $\delta=0.9$, and 15 stages with $\alpha_{\mathrm{LA}}=0.15$ and $\lambda_{\mathrm{BCL}}=0.1$.
\end{flushleft}
\end{table}

\textbf{Results on Places-LT.} 
Table~\ref{tab:placeslt} presents the evaluation results on the scene-centric Places-LT dataset using the ResNet-152 backbone. 
Places-LT is particularly challenging because scene categories exhibit severe imbalance and high visual ambiguity. 
Under this setting, \method remains highly competitive with strong decoupling, contrastive, and multi-expert methods. 
The fully integrated \method (BSM/BCL/LA) model achieves 42.15\% overall accuracy, outperforming standard decoupling baselines such as cRT (36.7\%) and LWS (37.6\%), as well as multi-expert methods such as RIDE (40.3\%) and SADE (40.9\%). 
More importantly, \method achieves strong Few-shot performance, reaching 40.82\% with BSM/BCL/LA. 
This compares favorably with strong ensemble and contrastive learning methods, including SADE+RL (38.7\%) and GPaCo+ConCutMix (34.9\%). 
Even compared with models trained using stronger augmentations, such as MDCS$^\dagger$ (36.3\% on Few-shot), \method maintains a clear advantage on the tail classes. 
These results suggest that the proposed trust-weighted ensemble is effective not only for object-centric datasets such as CIFAR-100-LT and ImageNet-LT, but also for scene recognition under severe class imbalance.

\textbf{Algorithm-level Interpretation.} 
It is important to note that no single long-tailed recognition algorithm uniformly dominates all datasets and all shot groups. 
Existing methods often exhibit different strengths: decoupled learning methods stabilize classifier learning, contrastive learning methods improve representation quality, and multi-expert methods enhance prediction diversity. 
\method follows a different design principle by serving as a modular ensemble framework that can absorb these complementary advantages. 
Rather than relying on a single re-balancing mechanism, \method constructs experts under different imbalance regimes and aggregates them according to class-wise reliability.

This property makes \method particularly effective for minority-class recognition. 
Across the reported comparisons, \method shows its clearest advantage on the Few-shot split for CIFAR-100-LT and Places-LT, where long-tailed classifiers typically suffer severe degradation. 
On CIFAR-100-LT, \method (BSM/BCL/LA) substantially improves Few-shot accuracy over strong baselines such as SADE, ConCutMix, and MGS. 
On ImageNet-LT, \method variants remain highly competitive with recent contrastive and ensemble methods while providing strong tail-class performance. 
On Places-LT, \method achieves one of the strongest Few-shot results despite severe scene-level imbalance. 
These results indicate that \method may not always maximize every individual metric, but it provides a robust and extensible algorithmic framework whose main advantage lies in improving underrepresented classes without severely sacrificing overall performance.

\begin{table}[htbp]
\centering
\small
\setlength{\tabcolsep}{4pt}
\caption{Top-1 accuracy (\%) comparison on Places-LT using ResNet-152.}
\label{tab:placeslt}
\begin{tabular}{lcccc}
\toprule
\textbf{Method}
& \textbf{Many} & \textbf{Medium} & \textbf{Few} & \textbf{All} \\
\midrule
\multicolumn{5}{l}{\textit{Standard long-tailed baselines}} \\
\midrule
Softmax (CE)
& 46.2 & 27.5 & 12.7 & 31.4 \\

Focal Loss~\cite{lin2017focal}
& 41.1 & 34.8 & 22.4 & 34.6 \\

$\tau$-norm~\cite{kang2020decoupling}
& 37.8 & 40.7 & 31.8 & 37.9 \\

cRT~\cite{kang2020decoupling}
& 42.0 & 37.6 & 24.9 & 36.7 \\

LWS~\cite{kang2020decoupling}
& 40.6 & 39.1 & 28.6 & 37.6 \\

LADE~\cite{hong2021disentangling}
& 42.6 & 39.4 & 32.3 & 39.2 \\

DisAlign~\cite{zhang2021disalign}
& 40.4 & 42.4 & 30.1 & 39.3 \\

Balanced Softmax~\cite{ren2020balanced}
& 42.6 & 39.8 & 32.7 & 39.4 \\

MiSLAS~\cite{zhong2021calibration}
& 39.6 & 43.3 & 36.1 & 40.4 \\

\midrule
\multicolumn{5}{l}{\textit{Multi-expert and recent methods}} \\
\midrule
BCL~\cite{zhu2022bcl,pan2024concutmix}
& 43.8 & 39.2 & 22.9 & 37.7 \\

BCL+ConCutMix~\cite{pan2024concutmix}
& 45.5 & 39.9 & 29.9 & 40.0 \\

RIDE~\cite{wang2021ride}
& 43.1 & 41.0 & 33.0 & 40.3 \\

ProCo+MORE~\cite{luo2025more}
& 43.3 & 42.2 & 33.1 & 40.8 \\

SADE~\cite{zhang2022sade}
& 40.4 & 43.2 & 36.8 & 40.9 \\

PaCo~\cite{cui2021paco}
& 37.5 & 47.2 & 33.9 & 41.2 \\

GPaCo~\cite{cui2024gpaco}
& 39.5 & 47.2 & 33.0 & 41.7 \\

PaCo+ConCutMix~\cite{cui2021paco,pan2024concutmix}
& 38.4 & 48.2 & 35.1 & 42.1 \\

GPaCo+ConCutMix~\cite{cui2024gpaco,pan2024concutmix}
& 39.9 & 47.8 & 34.9 & 42.2 \\

BLS+RL~\cite{ren2020balanced,zhao2024ltrl}
& 43.0 & 40.3 & 34.8 & 41.1 \\

LADE+RL~\cite{hong2021disentangling,zhao2024ltrl}
& 42.8 & 39.7 & 35.5 & 41.8 \\

RIDE+RL~\cite{wang2021ride,zhao2024ltrl}
& 43.1 & 41.9 & 36.9 & 42.1 \\

SADE+RL~\cite{zhang2022sade,zhao2024ltrl}
& 41.0 & 44.3 & 38.7 & 42.2 \\

ACL~\cite{ma2025acl}
& -- & -- & -- & 42.4 \\

\midrule
\textbf{\method} (BSM)
& 42.56 & 42.24 & 37.73 & 41.48  {\scriptsize $\pm$0.07} \\

\textbf{\method} (BSM/BCL)
& 43.05 & 42.90 & 38.34 & 42.07  {\scriptsize $\pm$0.06} \\

\textbf{\method} (BSM/LA)
& 40.95 & 42.69 & 40.05 & 41.55  {\scriptsize $\pm$0.06} \\

\textbf{\method} (BSM/BCL/LA)
& 41.34 & 43.39 & 40.82 & 42.15  {\scriptsize $\pm$0.08} \\

\midrule
\multicolumn{5}{l}{\textit{Stronger augmentation / alternate settings}} \\
\midrule
PaCo$^\dagger$~\cite{cui2021paco}
& 36.1 & 47.9 & 35.3 & 41.2 \\

NCL$^\dagger$~\cite{zhao2023mdcs}
& -- & -- & -- & 41.8 \\

MDCS$^\dagger$~\cite{zhao2023mdcs}
& 43.1 & 42.9 & 36.3 & 42.4 \\

PaCo+RL$^\dagger$~\cite{cui2021paco,zhao2024ltrl}
& 36.4 & 47.7 & 36.6 & 42.8 \\

\bottomrule
\end{tabular}
\vspace{0.3em}
\begin{flushleft}
\footnotesize

$^\dagger$ denotes methods trained with stronger augmentation or alternate settings, such as RandAugment.
Dashes indicate that shot-wise results were not reported in the corresponding paper.
\method uses $\tau=1.0$, decay rate $\delta=0.6$, and 15 stages with $\alpha_{\mathrm{LA}}=0.1$ and $\lambda_{\mathrm{BCL}}=1.2$.
\end{flushleft}
\end{table}

\subsection{Modular Integration of BSM, BCL, and LA}
\label{sec:bsm_bcl_la_synergy}

An important property of \method is that it is fundamentally an ensemble framework rather than a method restricted to a particular loss function. 
Therefore, it can flexibly incorporate advanced long-tailed learning components as long as they improve the quality or calibration of the stage-wise experts. 
In this work, BSM, BCL, and LA are combined because they play complementary roles in different parts of the pipeline. 
BSM is used as the main training objective and mitigates classifier bias toward head classes during expert optimization. 
BCL acts at the representation-learning level by encouraging more discriminative feature embeddings, which helps preserve separability for medium- and few-shot classes. 
LA is then applied as a post-hoc inference-time correction that adjusts residual class-frequency bias in the output logits without modifying the trained experts.

The empirical results support this complementary interpretation. 
On CIFAR-100-LT, the unified \method (BSM/BCL/LA) variant achieves 41.25\% Few-shot accuracy, showing a clear advantage over strong recent baselines. 
On ImageNet-LT, \method (BSM/BCL/LA) reaches 58.43\% overall accuracy, while the BSM/LA variant achieves 43.48\% Few-shot accuracy, indicating that inference-time calibration remains useful even after balanced training. 
On Places-LT, \method (BSM/BCL/LA) achieves 42.15\% overall accuracy and 40.82\% Few-shot accuracy, suggesting that the same combination remains effective under more severe scene-level imbalance.

These results highlight the modularity of \method. 
Structured sampling first produces diverse experts specialized for different imbalance regimes. 
BSM improves the class-balanced training of each expert, BCL strengthens the feature representations used by those experts, and LA refines their logits at inference time. 
The class-wise trust-weighted aggregation then combines these calibrated experts according to their estimated reliability for each class. 
Thus, the role of BSM, BCL, and LA is not to replace \method, but to provide stronger and better-calibrated experts for the proposed ensemble mechanism.

\section{Ablation Studies}
\label{sec:ablation}

\subsection{Effect of Ensemble Stages}
\label{sec:ablation_stage}

We first analyze how the number of ensemble stages affects \method.
Stage $m$ denotes an ensemble constructed by aggregating all experts from stage $1$ to stage $m$.
Thus, increasing $m$ corresponds to adding more experts trained on progressively clipped sub-training sets with different imbalance levels.
Early stages preserve the original long-tailed distribution more strongly, whereas later stages are trained on increasingly balanced subsets.
As a result, increasing the number of stages allows the ensemble to combine experts with complementary strengths across head and tail classes.
We evaluate CIFAR-100-LT, ImageNet-LT, and Places-LT using BSM loss with $\tau=2.0$.

\begin{figure}[t]
  \centering
  \includegraphics[width=0.8\linewidth]{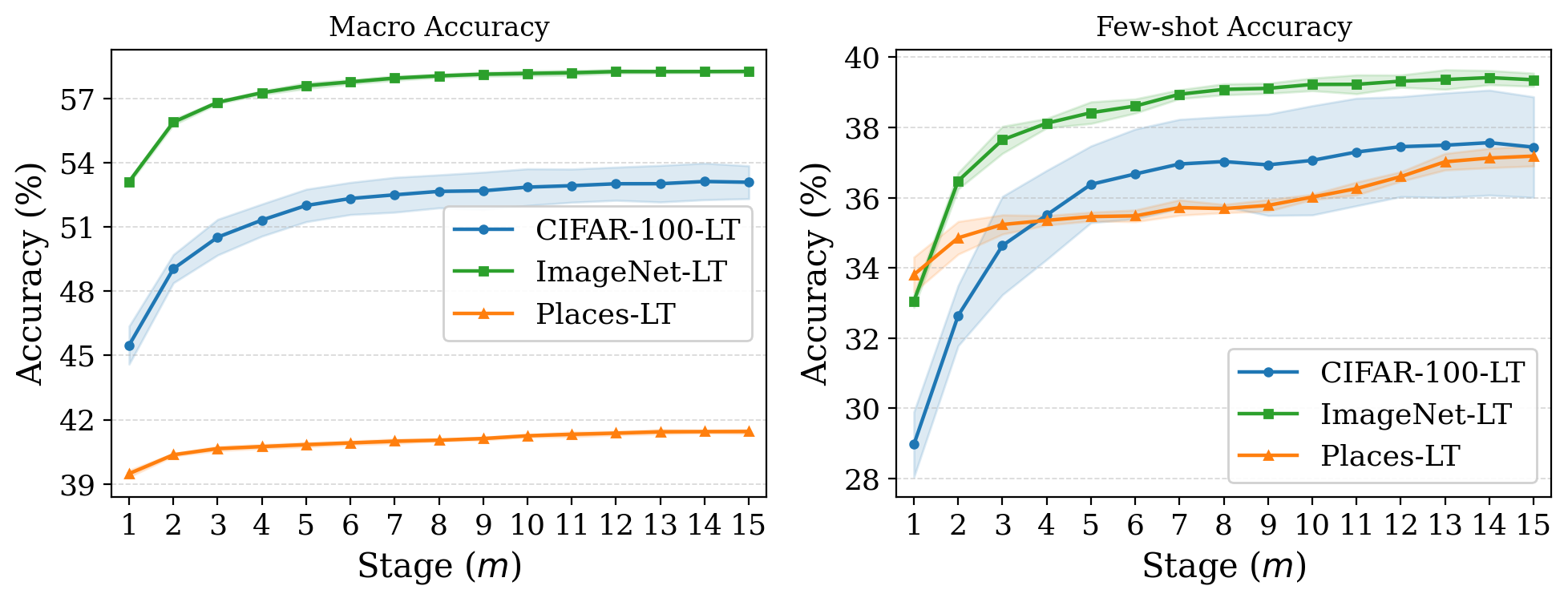}
  \caption{%
    \textbf{Macro Accuracy and Few-shot Accuracy vs.\ ensemble stage $m$.}
    All results use BSM loss with $\tau=2.0$.
    Shaded regions denote $\pm1$ standard deviation across seeds.
  }
  \label{fig:stage_curve_metric}
\end{figure}

Figure~\ref{fig:stage_curve_metric} shows that most gains are obtained in the early ensemble stages.
Macro Accuracy increases rapidly during the first few stages and then gradually saturates across all datasets.
This indicates that a small number of experts already captures the major distributional differences induced by structured sampling.
Few-shot Accuracy continues to benefit from additional stages, suggesting that later, more balanced experts remain useful for rare classes even after the overall performance begins to plateau.

The saturation after the early stages also suggests that the benefit of adding experts is not simply due to increasing ensemble size.
Instead, performance improves when newly added experts provide sufficiently different and useful class-wise behavior.
Once the clipped distributions become similar or the added experts no longer provide substantial complementary information, the marginal gain becomes small.
These results indicate that \method obtains most of its improvement from a compact set of early-stage experts.
In practice, 5--8 stages provide a reasonable trade-off between accuracy and inference cost.

\subsection{Effect of the Clipping Strategy}
\label{sec:clipping_ablation}

\begin{table}[t]
\caption{Top-1 accuracy (\%) performance on CIFAR-100-LT with different clipping methods using ResNet-32 experts.}
\label{tab:clipping_methods}
\centering
\small
\begin{tabular}{l|c|cccc}
\toprule
\textbf{Clipping Method} 
& \textbf{\# Experts} 
& \textbf{Many} 
& \textbf{Med.} 
& \textbf{Few} 
& \textbf{All} \\
\midrule
EDC ($\delta=0.9$) 
& 12 
& 68.3 & 51.4 & 35.1 & 52.4 {\tiny$\pm$0.71} \\

Uniform Interval 
& 12 
& 67.6 & 52.2 & 34.7 & 52.3 {\tiny$\pm$0.79} \\

Quantile Selection 
& 5 
& 66.2 & 50.4 & 32.4 & 50.5 {\tiny$\pm$0.85} \\

K-means Clustering 
& 7 
& 68.2 & 50.9 & 34.2 & 51.9 {\tiny$\pm$0.57} \\
\bottomrule
\end{tabular}
\end{table}

Table~\ref{tab:clipping_methods} compares different threshold-generation strategies on CIFAR-100-LT.
EDC achieves the best overall accuracy, while uniform interval selection performs competitively with the same number of experts.
In contrast, quantile selection and K-means clustering yield lower overall accuracy despite requiring fewer experts.
This suggests that overly adaptive threshold schedules may produce less stable expert sequences under long-tailed class distributions.
Overall, the smooth decay used by EDC provides a reliable way to generate progressively diversified experts, supporting the structured sampling design of \method.

We further analyze the effect of the EDC decay rate $\delta$, which controls how quickly the clipping threshold decreases across stages.
A smaller $\delta$ produces more aggressively clipped experts, increasing stage diversity but also reducing the amount of training data available to later experts.
We evaluate CIFAR-100-LT using BSM loss at the final ensemble stage and compare two sharpness settings, $\tau=2.0$ and $\tau=0.1$.

\begin{figure}[t]
  \centering
  \includegraphics[width=0.5\linewidth]{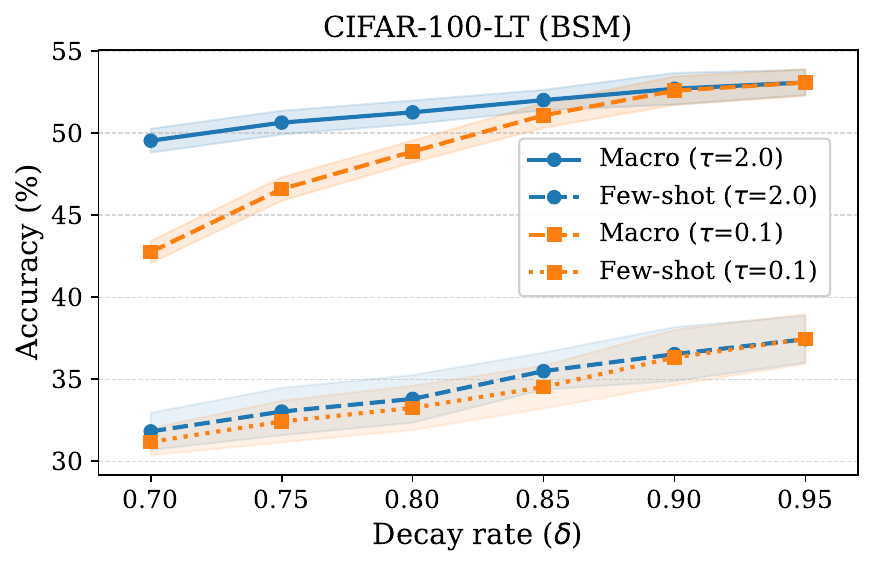}
  \caption{%
    \textbf{Macro Accuracy and Few-shot Accuracy vs.\ decay rate $\delta$ on CIFAR-100-LT.}
    All results use BSM loss.
  }
  \label{fig:dec_curve}
\end{figure}

Figure~\ref{fig:dec_curve} shows that larger values of $\delta$ generally improve both Macro Accuracy and Few-shot Accuracy.
This indicates that overly aggressive clipping may reduce expert quality, even if it increases diversity among stages.
The comparison between $\tau=2.0$ and $\tau=0.1$ further shows that sharper aggregation is more useful when $\delta$ is small, because aggressive clipping creates larger quality differences among experts.
As $\delta$ increases, the two sharpness settings become more similar, suggesting that the experts have comparable reliability.

Overall, $\delta$ controls the trade-off between expert diversity and expert reliability.
A stable clipping schedule preserves enough training data for later experts, while sharper class-wise weighting is mainly beneficial when the clipping process creates heterogeneous experts.

\subsection{Effect of the Ensemble Sharpness $\tau$}
\label{sec:tau_ablation}

We next analyze the effect of the ensemble sharpness parameter $\tau$.
Recall that $\tau$ controls how strongly \method favors high-trust experts in the class-wise aggregation.
When $\tau=0$, all experts contribute uniformly for each class, whereas larger $\tau$ makes the class-wise weighting more selective toward experts with higher estimated reliability.

\begin{figure}[t]
  \centering
  \includegraphics[width=0.8\linewidth]{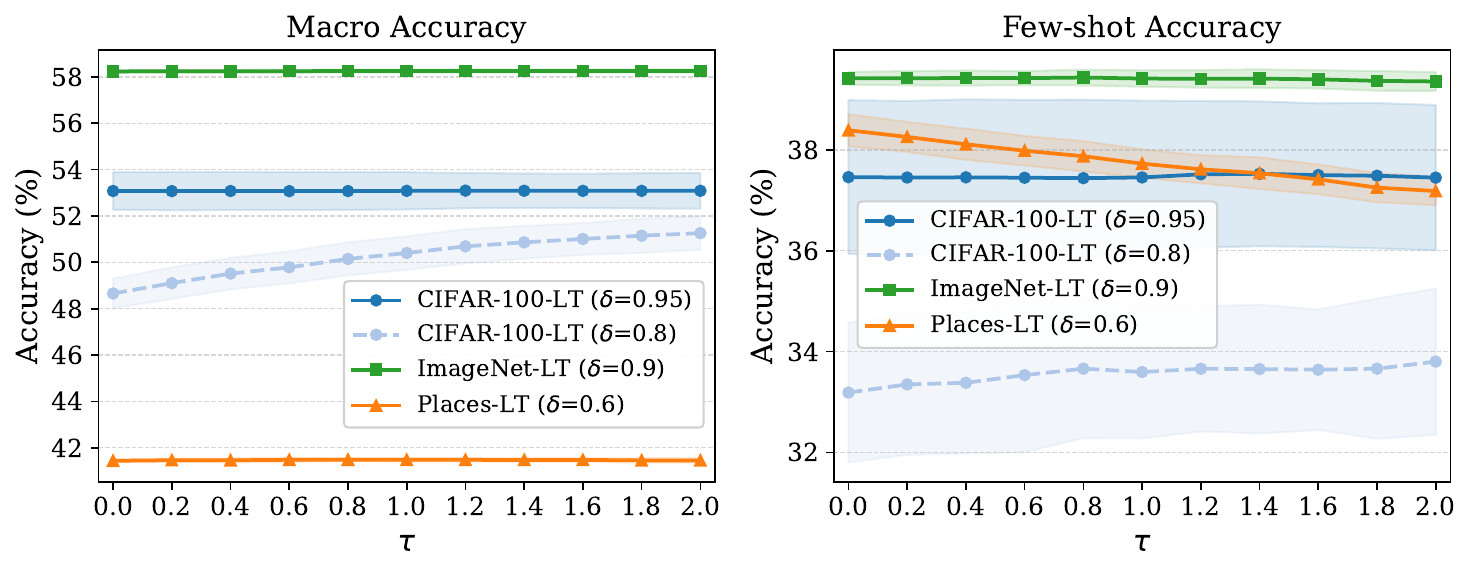}
  \caption{%
    \textbf{Macro Accuracy and Few-shot Accuracy as a function of $\tau$.}
    All results use BSM loss at the final ensemble stage.
  }
  \label{fig:tau_curve}
  \vspace{-2mm}
\end{figure}

Figure~\ref{fig:tau_curve} shows that the effect of $\tau$ depends on the degree of heterogeneity among experts.
When the clipping schedule is mild, the experts tend to have similar reliability, and both Macro Accuracy and Few-shot Accuracy remain relatively stable across different values of $\tau$.
In contrast, under a more aggressive clipping schedule, increasing $\tau$ improves performance by assigning lower weights to less reliable experts.
This suggests that sharper class-wise weighting is useful when the structured sampling process creates sufficiently diverse experts.

Places-LT exhibits a different pattern.
Although Macro Accuracy remains nearly stable, Few-shot Accuracy decreases as $\tau$ increases.
This suggests that class-wise reliability estimates for rare classes can be noisy when the available class-wise evidence is limited.
A large $\tau$ may over-amplify such noisy estimates and assign excessive weight to stages that appear reliable by chance.

Overall, $\tau$ acts as a stage-quality amplifier in the class-wise ensemble.
It is most beneficial when the structured sampling process creates heterogeneous experts, but overly sharp weighting can be harmful when class-wise reliability estimates are noisy.
This supports the motivation of \method's class-wise aggregation while suggesting that moderate sharpness is preferable in practice.

\subsection{Sensitivity to Expert Configuration}
\label{sec:expert_config_ablation}

We further analyze how optional expert-level configurations affect \method.
Specifically, we examine the BCL contrastive loss weight $\lambda_{\mathrm{BCL}}$ used for training the base experts and the post-hoc logit adjustment weight $\alpha_{\mathrm{LA}}$ used at inference time.

\begin{figure}[t]
  \centering
  \includegraphics[width=0.5\linewidth]{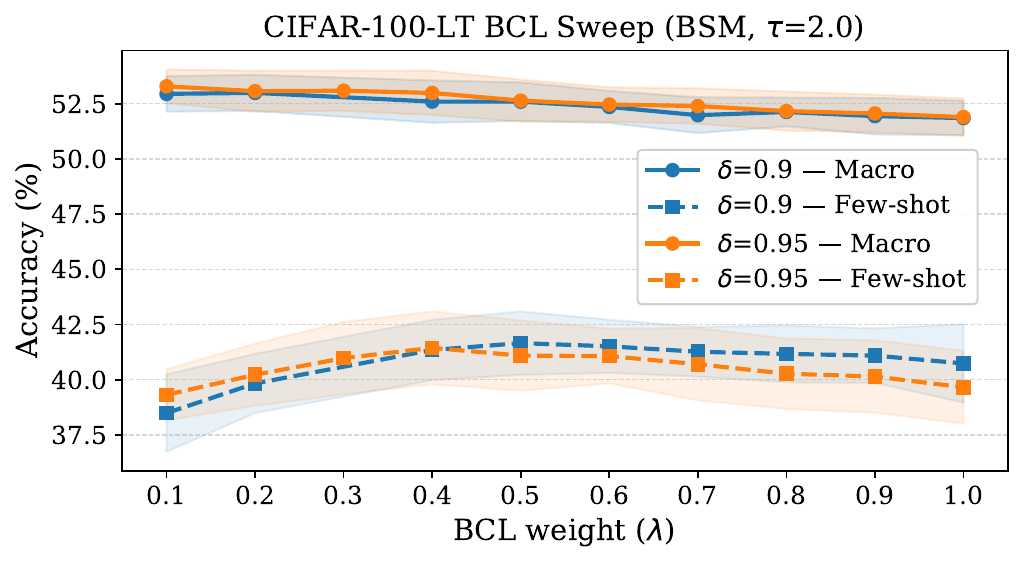}
  \caption{%
    \textbf{Effect of the BCL weight $\lambda_{\mathrm{BCL}}$ on CIFAR-100-LT.}
    BCL is used as the base expert objective within \method.
  }
  \label{fig:bcl_curve}
\end{figure}

\begin{figure}[t]
  \centering
  \includegraphics[width=0.8\linewidth]{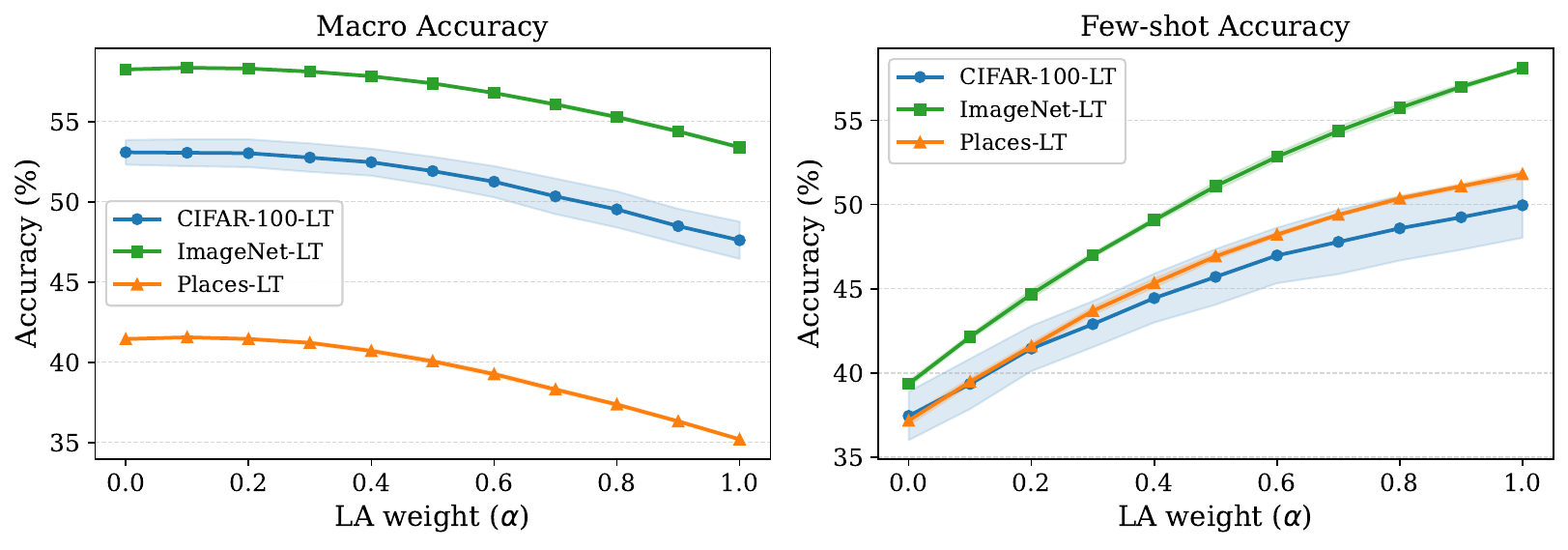}
  \caption{%
    \textbf{Effect of post-hoc logit adjustment $\alpha_{\mathrm{LA}}$.}
    LA is applied as an optional inference-time correction within \method.
  }
  \label{fig:la_curve}
\end{figure}

Figure~\ref{fig:bcl_curve} shows that $\lambda_{\mathrm{BCL}}$ mainly controls the trade-off between Macro Accuracy and Few-shot Accuracy.
A moderate BCL weight improves Few-shot Accuracy, while overly large values reduce Macro Accuracy.
Across the evaluated clipping schedules, the best Few-shot Accuracy is obtained around $\lambda_{\mathrm{BCL}}=0.4$--$0.5$, whereas smaller values such as $\lambda_{\mathrm{BCL}}=0.1$--$0.2$ better preserve Macro Accuracy.
This suggests that BCL can improve tail-class expert representations, but should be regarded as an interchangeable expert-training objective within \method.
Figure~\ref{fig:la_curve} shows that $\alpha_{\mathrm{LA}}$ provides an optional calibration knob for shifting the ensemble toward tail-class predictions.
Increasing $\alpha_{\mathrm{LA}}$ consistently improves Few-shot Accuracy across datasets, but excessively strong adjustment can reduce Macro Accuracy by over-correcting frequent classes.
In practice, small values such as $\alpha_{\mathrm{LA}}=0.1$--$0.2$ provide a conservative trade-off, while larger values are useful only when tail-class accuracy is prioritized.

Overall, these results show that expert-training and post-hoc calibration choices can adjust the operating point of \method, but they are orthogonal to the proposed class-wise trust-weighted aggregation.

\section{Limitations}

\method provides a flexible framework that combines diverse expert generation with class-wise trust-weighted aggregation.
Its main limitation is the additional computational cost introduced by training and aggregating multiple experts.
Although increasing the number of stages or using more diverse imbalance regimes can improve performance, it also increases training and inference overhead.
Moreover, overly aggressive clipping schedules may generate highly specialized or redundant experts whose marginal contribution is limited.
Therefore, the stage schedule and the number of experts should be chosen according to the dataset distribution and the available computational budget.

Another limitation concerns class-wise trust estimation for classes with $n_c \leq T_m$.
Because all available samples from these classes are retained in the corresponding sub-training set, no separate held-out samples remain, and trust is therefore estimated from training-side (in-bag) predictions.
This may lead to optimistic trust estimates, particularly for rare classes.
Although Beta-prior smoothing reduces overly extreme estimates when the number of relevant predictions is small, it does not fully eliminate this potential bias.

These limitations also open several directions for future work.
Since \method is modular, the overhead can be reduced by expert pruning, early stopping of uninformative stages, or parameter-efficient adaptation methods such as LoRA.
For example, a shared backbone with lightweight expert-specific adapters could replace fully independent expert models.
In addition, although our experiments mainly use BSM-based experts with optional LA or BCL, \method is not limited to these components.
The framework can incorporate other losses, sampling strategies, architectures, or expert types, as long as they provide complementary class-wise behavior.
Future work will explore automatic expert selection, more diverse expert generation strategies, and noise-resilient class-wise trust estimation.

\section{Conclusion}
This paper addressed a central challenge in long-tailed recognition: not merely low average accuracy, but uneven prediction reliability across different regions of the label space. 
To address this issue, we presented \method, a modular class-wise reliability-aware ensemble framework for long-tailed classification. 
Rather than treating each expert as uniformly reliable across all classes, \method estimates expert reliability at the class level and uses these estimates to guide ensemble aggregation.

\method combines threshold-based structured sampling with class-wise trust estimation. 
Structured sampling generates experts under different imbalance regimes while preserving the full label space, encouraging complementary behavior across head, medium, and tail classes. 
The resulting trust estimates are used in a class-wise generalized product-of-experts aggregation rule, allowing different experts to be emphasized for different predicted classes. 
This places class-wise expert reliability at the center of ensemble aggregation, rather than relying on a single global expert weight.

Experiments on CIFAR-100-LT, ImageNet-LT, and Places-LT show that \method achieves competitive overall accuracy and strong performance on underrepresented classes, with its clearest few-shot gains on CIFAR-100-LT and Places-LT while remaining competitive on ImageNet-LT. 
These findings support the value of modeling expert reliability at the class level, rather than relying only on global ensemble weights or average accuracy.

At the same time, \method introduces additional computational overhead because it relies on multiple experts. 
The ablation studies also suggest that overly aggressive stage schedules or sharp trust weighting can produce redundant experts or noisy trust estimates for extreme tail classes. 
Future work will investigate more efficient and robust expert construction strategies, including expert pruning, adaptive stage selection, uncertainty-aware trust estimation, and parameter-efficient expert adaptation. 
More broadly, \method suggests that class-wise expert reliability is a promising design principle for imbalanced classification and other knowledge-intensive settings where rare categories can be especially consequential.

\bibliographystyle{plainnat}
\bibliography{imbalance}

\end{document}